\documentclass[a4paper]{article}
\usepackage{graphicx}
\usepackage{multirow}

\usepackage{INTERSPEECH2020}
\usepackage[utf8]{inputenc}
\usepackage{tikz,booktabs,float,color,adjustbox}

\title{Efficient Weight factorization for Multilingual Speech Recognition}

\name{Ngoc-Quan Pham$^1$~~Tuan-Nam Nguyen$^1$~~Sebastian Stueker$^1$~~Alex Waibel$^1$$^,$$^2$}
\address{
  $^1$Interactive Systems Lab, Karlsruhe Institute of Technology, Karlsruhe, Germany\\
  $^2$Carnegie Mellon University, Pittsburgh PA, USA}
\email{ngoc.pham@kit.edu}

\begin{document}
\maketitle
\begin{abstract}
End-to-end multilingual speech recognition involves using a single model training on a compositional speech corpus including many languages, resulting in a single neural network to handle transcribing different languages. Due to the fact that each language in the training data has different characteristics, the shared network may struggle to optimize for all various languages simultaneously. In this paper we propose a novel multilingual architecture that targets the core operation in neural networks: linear transformation functions. The key idea of the method is to assign fast weight matrices for each language by decomposing each weight matrix into a shared component and a language dependent component. The latter is then factorized into vectors using rank-1 assumptions to reduce the number of parameters per language. This efficient factorization scheme is proved to be effective in two multilingual settings with $7$ and $27$ languages, reducing the word error rates by $26\%$ and $27\%$ rel. for two popular architectures LSTM and Transformer, respectively. 

\end{abstract}


\section{Introduction}

Multilingual modeling has been an important topic in applying sequence-to-squence models to language applications ranging from machine translation~\cite{ha2016toward,johnson2017google} to automatic speech recognition (ASR)~\cite{Waibel2000}. It is possible to employ one single neural model for multiple datasets with different languages with the goal of capturing the shared features between the languages. This method has been widely used to help under-resourced languages benefiting from the knowledge acquired from the richer counterparts. 


It is noticeable that the recent multilingual neural models are based on a semi-shared mechanism in which the largest body of the network architecture is exposed to all languages, while a smaller weight subset provides a language specific bias. This was shown to be more effective in a multilingual scenario than fully sharing the whole network\cite{ha2016toward,johnson2017google} since each language has certain unique features, and the single architecture often struggles to handle a variety of languages~\cite{platanios-etal-2018-contextual}.  

There are two main drawbacks that are typically presented in the existing implementations of the semi-shared mechanism. On the one hand, the implementations often depends heavily on a certain architecture being popular at the time, and the given improvement is going to be diminished when a new architecture evolves. For example, the language-specifically biased attention~\cite{zhu2020multilingual} modified the self-attention architecture~\cite{vaswani2017attention} specifically based on the assumption that each language can benefit from a bias added to the attention scores. On the other hand, the language-dependent components might require a considerable amount of parameters and struggles to scale to the number of languages. For example, the language adapters added to the Transformer layers~\cite{bapna2019simple} are essentially feed-forward neural network layers being similar to the counterpart already in the shared Transformer body. A scenario with $20$ languages consequently generates hundreds of these layers accounting for a large amount of parameters to be optimized.   

In this work, we propose a multilingual architecture using a factorization scheme that is both effective and highly scalable with the number of languages involved. Moreover, this scheme is applicable to any neural architectures as long as matrix-vector multiplication is the dominant operation. The key idea of our work is that each weight matrix in the shared architecture can be factorized into a shared component and multiple additive and multiplicative language dependent components. While each language is assigned with extra weights to learn distinctive features, simplicity and scalability are achieved by further representing those weights into as a rank-1 matrix, thus can be factored into two vectors. This method is demonstrated to be computational friendly with a minimal overhead and can be applied to a arbitrary neural architecture.

Subsequently, this weight factorization method is then evaluated on two different scenarios: one with $7$ languages having similar amounts of data, and one with $27$ languages with various extremely low resource data. The method is implemented on two commonly used architectures: Long Short-Term Memories (LSTM) and Transformers which show that both types of networks can benefit by weight factorization in multilingual ASR. The reduction of error rate can be up to $47\%$ rel. in the case of low-resource languages such as Japanese\footnote{Error is measured in characters error rate here.} and  $15.5\%$ rel. on average with the moderately sized languages.

\section{Methodology}

A neural speech-to-text model  transforms a source speech input with $N$ frames $X = {x_1, x_2, \dots, x_N}$ into a target text sequence with $M$ tokens $Y = {y_1, y_2, \dots, y_M}$. The encoder transforms the speech input into higher level feature vectors $h^X_{1\dots N}$. The decoder jointly learns to generate the output distribution $o_i$ based on the previous target tokens ${y_1, y_2, \dots, y_{i-1}}$ while looking for the relevant inputs from the input via the attention mechanism~\cite{bahdanau2014neural,vaswani2017attention}. 

\begin{align}
    h^X_{1\dots N} &= ENCODER(x_1 \dots x_N) \\
    h^Y_i &= DECODER(y_i, y_{1 \dots {i-1}})  \\
    c_i &= \text{ATTENTION}(h^Y_i, h_{1\dots N}) \\
    o_i &= \text{SOFTMAX}(c_i + h^Y_i) \label{eq:combine} \\ 
    y_{i+1} &= sample(o_i) 
\end{align}

Notably, there is a large variety of model architectures that implement this encoder-decoder design. The core networks in the encoder and decoder range from LSTM~\cite{bahdanau2016end}, convolution/TDNN~\cite{gehring2017convolutional,zhang2017very} to self-attention~\cite{pham2019transformer} or even a mix of the above~\cite{gulati2020conformer}

The universal multilingual framework~\cite{ha2016toward,johnson2017google} employs a single model to learn on a joint training dataset containing multiple languages, which is different than the predating multi-way encoder-decoder approach~\cite{firat2016multi}. 

\subsection{Multilingual weight composition}

It can be seen that, the common ground of the aforementioned architectures is the usage of linear combinations of lower level features $X \in R^D$ which can be expressed as the matrix multiplication between input $X$ and a weight matrix $W$. For example, the LSTM contains four different projections for its forget, input, output gates and candidate content~\cite{hochreiter1997long}, as can be seen in Equation~\ref{eq:lstm}.

\begin{align}
\label{eq:lstm}
    f_t = sigmoid(W_{fx}^\top X_t + W_{fh}^\top H_{t-1} + b_f) \\
    i_t = sigmoid(W_{ix}^\top X_t + W_{ih}^\top H_{t-1} + b_i) \\
    \hat{c}_t = tanh(W_{cx}^\top X_t + W_{ch}^\top H_{t-1} + b_c) \\
    o_t = sigmoid(W_{ox}^\top X_t + W_{oh}^\top H_{t-1} + b_o) 
\end{align}

Similarly, the main components of the Transformer layers are self-attention layers and feed-forward layers. While the latter are fundamentally two layers of linear projections, the former is also comprised of linear projections that generate queries $Q$, keys $K$ and values $V$ from the input $X$:

\begin{align}
\label{eq:attention}
    Q = W_Q^\top X \\ 
    K = W_K^\top X \\
    V = W_V^\top X \\
    SelfAtt(X) = softmax(QK^\top)V 
\end{align}

The main idea here is that each matrix multiplication $Y = W^TX$ in the multilingual model can be decomposed into a function of shared weights $W_S$ and additional language dependent weights $W_{ML}$ and $W_{BL}$

\begin{align}
\label{eq:decompose}
    Y = (W_S \cdot W_{ML} + W_{BL})^\top X \\
\label{eq:decompose2}
      = (W_S \cdot W_{ML})^\top X +  W_{BL}^\top X
\end{align}

Here the added weights include the first multiplicative term $W_{ML}$ that directly change the magnitude and direction of the shared weights $W_S$ and the  biased term $W_{BL}$ provides the network with a content-based bias depending on the input features $X$. Each language maintains a distinctive set of $W_{ML}$ and $W_{BL}$ so that the whole architecture is semi-shared.


\subsection{Factorization}

There is, however, an obstacle that both $W_{ML}$ and $W_{BL}$ require to be the same size with $W_S$, which makes the language dependent weights dominate the shared weights, while the intuition is the opposite. Fortunately, it is possible to use rank-1 matrices $\bar{W} \in R^{D_{in} \times D_{out}}$ that can be factorized into vectors~\cite{wen2020batchensemble,yoon2019scalable}, for example with two vectors $r \in R^{D_{in}}$ and $s \in R^{D_{out}}$ such that $\bar{W} = r s^\top$ which reduces the number of parameters from $D_{in} \times D_{out}$ to $D_{in} + D_{out}$. 

One drawback in this method is the lacking representational power of Rank-1 matrices. One solution is to modify the factorization into using $k$ vectors per language so that there are $k$ independent weight factors followed by a summation, which increases the rank of the additional weight matrices.

\begin{equation}
    \bar{W} = \sum_i^k r_i s_i^\top
\end{equation}


\subsection{Computational cost}

The factorization above is applied to both $W_{ML}$ and $W_{BL}$ to ensure that the dominated force is still the shared weights, while each language at $k=1$ is characterized by an additional $ \frac{D_{in} + D_{out}}{D_{in} \times D_{out}}$ amount of weights. In a typical network architecture with $D_{in}$ and $D_{out}$ being typically $512-2048$, this amounts for $0.1-0.3$ percents of the total network's weights per language, therefore scalable to hundreds. 

On the time complexity, the amount of extra computation comes from generating the combinatory weight $W$ from $W_S$ and the multiplicative/bias terms $W_{ML}$ and $W_{BL}$. Fortunately, this  overhead coming from element-wise multiplication and addition is rather small compared to the matrix multiplication. More importantly, it is possible to utilize the optimized implementation of the original network\footnote{such as the CUDA implementations of LSTM and Self-Attention} which minimizes the computational requirements of our approach. 

On the same subject,~\cite{wen2020batchensemble} proved that $W$ does not have to be explicitly computed, but their approach required to rewrite the graph operation for the core networks in popular deep learning frameworks. 

\section{Related works and Comparison}

In the world of speech recognition, training a single recognizer for multiple languages is not a thematic stranger~\cite{Waibel2000} from Hidden Markov Model (HMM) based models~\cite{burget2010multilingual,lin2009study}, hybrid models~\cite{heigold2013multilingual} to end-to-end neural based models with CTC~\cite{muller2018neural,kim2018towards} or sequence-to-sequence models~\cite{toshniwal2018multilingual,zhu2020multilingual,zhou2018multilingual,adams-etal-2019-massively,kannan2019large,li2019bytes}, with the last approach being inspired by the success of multilingual machine translation~\cite{ha2016toward,johnson2017google}. The literature especially mentions the merits of disclosing the language identity (when the utterance is supposed to belong to a single language) to the model, whose architecture is designed to incorporate the language information. 

One of the manifestations is language gating from either language embeddings~\cite{kim2018towards} or language codes~\cite{muller2018neural,muller2019neural} that aim at selecting a subset of the neurons in the network hidden layer. In our current approach, this effect can be achieved by factorizing further Equation~\ref{eq:decompose2}~\cite{wen2020batchensemble}:
\begin{align}
\label{eq:factorize2}
    Y = (W_S \cdot W_{ML})^\top X +  W_{BL}^\top X \\
      = (W_S \cdot (r_m s_m^\top)^\top X + (r_a s_a^\top)^\top X \\
\label{eq:factorize3}
      = (W_S^\top (X \cdot s_m) \cdot r_m) +  (r_a s_a^\top)^\top X
\end{align}

In Equation~\ref{eq:factorize2}, the multiplicative matrix $W_{ML}$ is factorized by two vectors $r_m$ and $s_m$. The left hand size of Equation~\ref{eq:factorize3} shows us that the those vectors can be learned to gate the input vector $X$ and the output of the linear projection  $(W_S^\top (X \cdot s_m)$. This intuition also suggested us to initialize $r_m$ and $s_m$ to one-vectors similarly to normalization techniques~\cite{ioffe2015batch,ba2016layer}. Since layer normalization often comes before the linear projection layers in Transformers, this scheme also helps our model to generalize to assigning to each language a different normalization scale and variance~\cite{zhang-etal-2020-improving}. 

On the other hand, the right hand side of Equation~\ref{eq:factorize3} gives us the bias to the linear projection which has been used in either language embeddings~\cite{pratap2020massively} and customized attention layers with language biases~\cite{zhu2020multilingual}. 

A different line of research involves using language code~\cite{muller2018neural} to differentiate language coming from a separate classifier. The language code is often trained separately and then mixed into the ASR architecture later~\cite{muller2019neural} giving the lingual bias. Our method can provide a similar effect with end-to-end training and without architectural modification. The advantage of this method is to exploit unlabeled (transcript-wise) data to gather language-specific information.  

Architecture wise,~\cite{bapna2019simple} makes the network language aware using language-dependently adaptive feed-forward layers at the end of each Transformer block. While this method is able to be effective in translation~\cite{philip2020language} and speech recognition scenarios~\cite{zhu2020multilingual}, it requires a considerable amount of parameters per language\footnote{Each feed-forward component accounts for around 25\% the amount of parameters of each encoder block.} and probably becomes incompatible with future architectures because it is specifically designed for Transformers. 

The closest to our work is the parameter generator~\cite{platanios-etal-2018-contextual} that composes a weight matrix $W \in R^{D_{in} \times D_{out}}$ using a shared tensor $W_S \in R^{D_{in} \times D_{out} \times D_L}$ and a language embedding vector $L \in R^{D_L}$. The main disadvantage with that approach is that the amount of parameters linearly scales in the size of the language embedding $D_L$, and the whole body of parameters participates in every language. Our initial experiments cannot produce a reasonable result for a straight comparison, partly because the memory is quickly overwhelmed by the number of parameters. 

For a larger context, weight factorization has been investigated to generate distinguishable yet cheaper copies of an existing network to allow for economical ensembles~\cite{wen2020batchensemble}, Bayesian networks~\cite{dusenberry2020efficient} or continual learning without catastrophe forgetting~\cite{yoon2019scalable}. Similar ideas to use different weights for different languages have been investigated early on by~\cite{hampshire1992meta}. 

\section{Experiments}

\subsection{Datasets}
The effects of the weight factorization methods are measured on datasets publicly available including Mozilla Common Voice~\cite{ardila2019common} containing up to 27 languages, Euronews~\cite{gretter2014euronews} and Europarl-ST~\cite{iranzo2020europarl} having $4$ and $9$ languages respectively. The preprocessing steps include converting audio into $40$-dimensional feature frames, and generating BPE for each language with $256$ codes each. Only Japanese and Chinese are handled at character level\footnote{Our initial experiments with joined BPE gave worse results for the 27-language dataset}. All of the three mentioned datasets come with the predefined validation and test partition, which are used in our experiments. 

Two experimental scenarios are investigated in our work: initially we work on a set of $7$ European languages: German (de), Italian (it), Spanish (es), Dutch (nl), French (fr), Polish (pl) and Portuguese (pt) each of which contain at least $60$ hours of training data. The second scenario later expands to a total of $27$ languages of more origin and diversity.


\subsection{Model and Training description}
The experiments are conducted with two model architectures, two of which are commonly used in end-to-end speech recognition~\cite{zeyer2019comparison}: a) LSTM-based encoder-decoder networks~\cite{park2019specaugment} in which the LSTMs have $1024$ hidden units and the encoder is downsampled using two $3\times3$-filter convolutional layers, and b) Transformer networks~\cite{vaswani2017attention} with relative attention~\cite{pham2020relative} with weight factorization for this multilingual setup. For the Transformer, we use the Transformer-Big configurations in~\cite{vaswani2017attention} with model size $1024$ but with $16$ encoder layers with stochastic layer dropout with the same setting as in~\cite{pham2019transformer}.

All models are trained on single GPU by grouping a maximum $45,000$ frames per mini-batch\footnote{speech inputs are often longer than their transcriptions, so grouping mini-batches by frames is more efficient}, and the gradients are updated every 16 mini-batches with adaptive scheduling in~\cite{vaswani2017attention} using the base learning rate $1.5$ and $4,096$ warm-up steps. The inputs are masked with SpecAugmentation~\cite{park2019specaugment}. Given the large configuration, we train all models up to $150,000$ updates or up to 2 weeks. It is notable that the factorized versions have minimal overhead which results in a 10 percent training speed reduction, while the adapter method requires at least $33\%$ more time.

\subsection{Baseline models}
The comparison in the upcoming result section involves two previous works that were re-implemented. First, the language embedding was concatenated to the speech features and word embeddings at the encoder and decoder respectively which was used in~\cite{pratap2020massively}. Second, the language dependent adapters~\cite{bapna2019simple} were used. In this case, we use adapters in the form of feed-forward networks with $1,024$ neurons in the hidden layer. While theoretically the language embedding is a subset of our factorized network because the former is essentially a small set of weights dedicated for each language, the adapter network is fundamentally different because it requires extra layers, adding depths and nonlinearity levels to the overall architecture, while our factorization scheme keeps the interaction between inputs and weights unchanged.  


\subsection{Experiments with 7 languages}

The word error rates for each language using two baseline models (with Transformer (TF) and LSTM), their factorized versions and the TF with adapter~\cite{bapna2019simple} are shown in Table~\ref{tab:result7}. Averaged over the 7 languages, the error rate is reduced by $15.5\%$ and $7.2\%$ rel. for the Transformer and LSTM respectively, and the improvement is significant across languages, unlike the Adapter technique which manages to reduce the error rate for 4 languages but is not better for the other languages. 

Regarding the number of parameters, the Transformer and its factorized variation has twice as many parameters as the LSTMs, thus possibly explaining the improvement regarding performance. While this seems to contradicts the large number of parameters for the ADT model that needs $42\%$ more space than the factorized TF, the ADT actually adds more depth (2 per TF block). This is a significant change to the architecture because with layer normalization, all languages share the same layer mean and variance at each level, while this is not changed with the adapter.

\subsection{Experiments with 27 languages}

Under this condition, the factorization method maintains the improvement across all languages, with overall $26\%$ rel. WER reduction in average for Transformer and $27.2\%$ rel. for LSTM, as summarized in Table~\ref{tab:result27}. Importantly, the factorized models are effective while using only $15\%$ more parameters, while the ADT Transformer needs almost 1 billion parameters to achieve a $21.2\%$ rel. improvement, due to each language requiring $2$ more layers per block. 

While the most resourceful languages such as German, Italian, Spanish and French observe the similar improvement compared to the 7-language experiment, the lower resource counterparts are often improved significantly compared to the baseline, regardless of the model architecture. The error rates on Japanese and Latvian testsets were decreased by $48\%$ rel. compared to the base Transformer, and multiple languages were improved by $30\%$ rel. including Arabic, Br, Cnh, Cv and Ta. The only language that remains a relative high error rate is Dhivehi, in this case staying over $60\%$ regardless of the architecture. One explanation for the large improvements regarding lower resource languages is that, the language weights are only learned to optimize for those particular languages, while the shared weights are frequently changed attempting to optimize all different language/task losses. This problem is often alleviated using learnable and weighted sampling~\cite{wang-etal-2020-balancing} to help the gradients remain stable for the less frequently visited languages. 

A direct comparison between two Transformer variations shows that the factorization is consistently better in $21$ languages and the adapters yielded better results in $6$, given the same time and computational constraints. While it is also possible for the adapters to obtain better performances by longer training, the presented results provide evidences that our proposed factorization scheme is able to outperform both the baseline and the deeper language adapter network without extensive tuning and with reasonable resources. 

\begin{table}[ht]
\caption{Comparison on the 7-language dataset (WER$\downarrow$). Our baseline models include the Transformers (TF), LSTM and their factorized (FTR) variations respectively. The last column is the Transformer with Adapter (ADT)~\cite{bapna2019simple}.}
\label{tab:result7}
	\centering
	\begin{tabular}{lccccc}
		\toprule
        \textbf{Language} & \textbf{TF} & \textbf{+FTR} & \textbf{LSTM} &  \textbf{+FTR} & \textbf{ADT} \\
        \midrule
        \# Params & 335M & 350M & 167M & 172M & 497M \\
        \midrule
        de & 15.78 & 14.62 & 15.75 & 15.53 & 14.71 \\   
        es & 16.06 & 13.47  & 14.66 & 14.09 & 14.81 \\
        fr & 17.34 & 16.26 & 17.35 & 16.44 & 16.76 \\
        it & 18.62 & 15.82 & 16.65 & 15.63 & 17.58 \\
        nl & 26.61 & 22.33 & 24.18 & 22.57 & 31.84 \\
        pl & 20.4  & 15.7  & 16.39 & 15.28 & 20.65 \\
        pt & 25.8  & 19.3 & 23.21 & 19.49 & 25.19 \\  
        \midrule
        Mean & 20.08 & \textbf{16.97} & 18.31 & 17.00 & 20.2 \\
		\bottomrule
	\end{tabular}
\end{table}

\begin{table}[ht]
\caption{Comparison on the 27-language dataset. The models being shown include Transformers (TF), LSTM (TF) and their factorized versions (FTR). WER$\downarrow$~.}
\label{tab:result27}
	\centering
	\begin{tabular}{lccccc}
		\toprule
        \textbf{Language} & \textbf{TF} & \textbf{+FTR} &  \textbf{LSTM} &  \textbf{+FTR} & \textbf{ADT} \\
        \midrule
        \# Params & 355M & 416M & 177M & 194M  & 980M \\
        \midrule
         (ar) &  26.2 & \textbf{17.81} & 28.73 & 20.02 & 16.56 \\
               (br) & 51.85 & \textbf{34.69} & 71.53 & 40.49 & 40.21 \\
               (cnh)& 52   & \textbf{38.33}  & 62.19 & 36.59 & 55.18 \\
               (cv) & 53.88 & \textbf{33.11} & 61.61 & 39.6 & 38.40 \\
         (de) & 16.89 & \textbf{15.62} & 19.89 & 16.59 & 16.35 \\
               (dv) & 71.63 &\textbf{ 63.72}	& 80.18	& 64.82 & 65.23 \\
               (es) & 16.05 & \textbf{14.53}	& 18.41	& 14.82 & 15.27 \\
       (et) & 33.95	& 30.43	& 39.63	& 34.26 & \textbf{28.12} \\
               (fr) & 18.61	& \textbf{17.24}	& 20.86	& 17.43 & 17.87 \\
               (ia) & 49.86	& 33.24	& 48.39	& \textbf{31.96} & 42.40\\
               (id) & 28.78	& \textbf{17.28}	& 32.9	& 20.22 & 22.79 \\
               (it) & 20.76 & \textbf{18}	& 21.99	& 18.07 & 19.60 \\
               (ja) & 39.17	& \textbf{20.44}	& 38.92	& 23.79 & 27.55 \\
               (lv) & 66.17	& \textbf{34.3}	& 66.66	& 37.93 & 43.57 \\
               (ky) & 22.08 & \textbf{17.17}	& 18.68	& 21.46 & 12.86 \\
               (mn) & 42.03	& \textbf{35.03}	& 46.42	& 38.5 &  34.12 \\
               (nl) & 27.54	& \textbf{23.75}	& 29.44	& 23.93 & 28.30 \\
               (pl) & 21.81	& 17.8	& 19.92	& \textbf{17.19} & 18.75 \\
               (pt) & 25.16	& 21.38	& 27.13	& \textbf{21.37} & 22.82 \\
               (ro) & 39.39	& 32.15	& 34.7	& \textbf{26.73} & 41.71 \\
               (sah)& 57.47	& 50.47	& 69.04	& \textbf{49.2} & 55.27 \\
               (sl) & 49.73	& 22.01	& 48.92	& 29.66 & \textbf{20.77} \\
               (ta) & 33.1	& 22.34	& 18.87	& 28 & \textbf{16.36} \\
               (tr) & 6.04	& 5.16	& 4.99	& 8.29 & \textbf{2.40} \\
               (tt) & 24.96 & 22.12	& 38.03	& 24.07 & \textbf{21.83} \\
               (zh) & 24.05	& \textbf{22.53} 	& 33.01	& 23.54 & 25.99\\
        \midrule
        Mean & 35.4 & \textbf{26.2} & 38.5 & 28.0 & 27.78 \\
		\bottomrule
	\end{tabular}
\end{table}


\section{Conclusion}
In this work, we proposed a method to decompose and factorize weights enabling multilingual end-to-end ASR models to learn more efficiently. While the main results are promising and the method can be applied to arbitrary neural architectures, we are also aware that method requires the utterance to contain a single language and thus is limited to such scenarios. Future work will investigate the usage in a code-mixing scenario and incorporating unlabeled data for language-specific feature learning.

\section{Acknowledgements}
We thank Jan Niehues for suggesting the similarity between the multiplicative weights and layer normalization. 

Parts of this work were realized within the project ELITR which has received funding from the European Unions Horizon 2020 Research and Innovation Programme under grant agreement No 825460.

Parts of this work were realized within a project funded by the Federal Ministry of Education and Research (BMBF) of Germany under the number 01IS18040A.

\bibliographystyle{IEEEtran}

\bibliography{mybib}
\newpage

\end{document}